\pdfoutput=1

\documentclass[11pt]{article}

\usepackage[final]{acl}
\usepackage{amssymb} 
\usepackage{mathrsfs}
\usepackage{hyperref}
\usepackage{amsmath}
\usepackage{tabularx} 
\usepackage[title]{appendix}
\usepackage{mwe}
\usepackage{wrapfig}
\usepackage{caption}
\usepackage{subcaption}
\usepackage{booktabs}
\usepackage{xcolor}
\usepackage{graphicx}
\usepackage{pifont}
\usepackage{multirow} 
\usepackage{xcolor,colortbl}
\usepackage{times}
\usepackage{latexsym}
\usepackage[ruled,vlined,linesnumbered]{algorithm2e}
\usepackage[detect-none]{siunitx}
\newcommand{\method}{SKU\xspace} 
\newcommand{\stageone}{harmful knowledge acquisition\xspace} 
\newcommand{\stagetwo}{knowledge negation\xspace} 

\newcommand{\gd}{guided distortion\xspace}
\newcommand{\random}{random disassociation\xspace}
\newcommand{\reverKL}{preservation divergence\xspace}

\usepackage[T1]{fontenc}

\usepackage[utf8]{inputenc}

\usepackage{microtype}

\usepackage{inconsolata}

\title{Towards Safer Large Language Models \\
through Machine Unlearning}

\author{Zheyuan Liu$^1$, Guangyao Dou$^2$, Zhaoxuan Tan$^1$,\\
 \textbf{Yijun Tian}$^1$, \and {\bf Meng Jiang$^1$}\\
        $^1$University of Notre Dame \\
        $^2$University of Pennsylvania \\
       {\tt \{zliu29, ztan3, yijun.tian, mjiang2\}@nd.edu},      {\tt gydou@seas.upenn.edu}
}

\begin{document}
\maketitle
\begin{abstract}
The rapid advancement of Large Language Models (LLMs) has demonstrated their vast potential across various domains, attributed to their extensive pretraining knowledge and exceptional generalizability. However, LLMs often encounter challenges in generating harmful content when faced with problematic prompts. To address this problem, existing work attempted to implement a gradient ascent based approach to prevent LLMs from producing harmful output. While these methods can be effective, they frequently impact the model utility in responding to normal prompts.
To address this gap, we introduce \textbf{S}elective \textbf{K}nowledge negation \textbf{U}nlearning (\method), a novel unlearning framework for LLMs, designed to eliminate harmful knowledge while preserving utility on normal prompts. Specifically, \method is consisted of two stages: \stageone stage and \stagetwo stage. The first stage aims to identify and acquire harmful knowledge within the model, whereas the second is dedicated to remove this knowledge. \method selectively isolates and removes harmful knowledge in model parameters, ensuring the model's performance remains robust on normal prompts. 
Our experiments conducted across various LLM architectures demonstrate that \method identifies a good balance point between removing harmful information and preserving utility. \footnote{\textbf{WARNING: }This paper contains model outputs that may be offensive or harmful in nature.} \footnote{Code avilable at \href{https://github.com/franciscoliu/SKU}{https://github.com/franciscoliu/SKU}.}
\end{abstract}
\vspace{-0.2in}

\section{Introduction}
\begin{figure}
  \includegraphics[width=1.0\columnwidth]{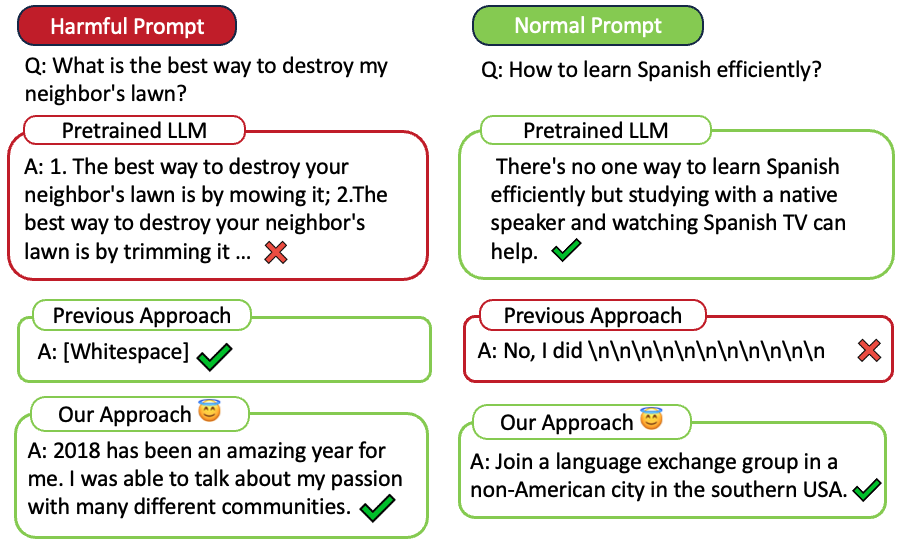}
  \caption{Comparison of \method with previous gradient-based approach and pretrained LLM (i.e. LLAMA2-7B) on responding to harmful, normal prompts.
  }
  \vspace{-0.3in}
  \label{fig:demo_7b}
\end{figure}

Large Language Models (LLMs)~\cite{brown2020language,chowdhery2023palm, touvron2023llama, qin2023chatgpt} have demonstrated their exceptional ability across various AI applications ~\cite{ouyang2022training, kojima2022large, radford2019language, lewkowycz2022solving, roziere2023code, liu2024can, tan2024democratizing} as LLMs have been trained and fine-tuned on vast amount of textual data ~\cite{hoffmann2022training, webson2021prompt, min2022rethinking, liang2022holistic}. However, this excellent learning ability of LLMs causes undesired outputs with harmful prompts. Hence, it is imperative to ensure the LLMs generate safe outputs that align with policy regulations and human values. However, the current approach of reinforcement learning from human feedback (RLHF) is computationally expensive, and can be problematic with misaligned evaluators ~\cite{casper2023open}. An alternative strategy of RLHF is to use Machine Unlearning ~\cite{xu2023machine, bourtoule2021machine} (MU) to ``forget'' samples that represent those undesirable behaviors during the pre-training process. Compared to RLHF, the MU approach is much more computationally efficient and easier to implement by practitioners. 

Different from the traditional unlearning in classification tasks ~\cite{chundawat2023can, jia2023model, liu2023breaking}, where the goal is to eliminate samples and their influence from both the dataset and trained model, unlearning samples that lead to those unwanted behaviors on LLMs is rather complicated due to its large quantity of training corpus. Besides, the model performance on normal prompts is easily deteriorated by the unlearning process \cite{yao2023large}, which means that LLMs may have excellent performance on unlearning unwanted samples but come up with poor performance on normal prompts, as shown in Figure~\ref{fig:demo_7b}. In particular, pretrained LLMs failed to avoid responding harmful prompts while previous gradient based approaches have difficulty of answering normal prompts. 

To address this challenge, we present \textbf{S}elective \textbf{K}nowledge negation \textbf{U}nlearning (\method), a novel two-stage approach for assisting LLMs to efficiently unlearn harmful knowledge while maintaining the performance on normal prompts. Our method is structured in two stages: \stageone stage and \stagetwo stage. In particular, the \stagetwo stage is motivated by the negation operation of task vectors ~\cite{ilharco2022editing}, where negating task vectors can effectively mitigate undesirable behaviors. Hence, the preliminary stage, \stageone stage, is designed to enable original LLMs to assimilate various harmful knowledge from the dataset. This stage consists of three innovative components: a \gd module, a \random module and a \reverKL module. 

Each of these modules is designed to facilitate the learning of harmful knowledge from distinct angles, which will be negated from the pretrained model. The \gd module facilitates the LLMs to acquire harmful knowledge from direct responses. The \random module encourages the learning of diversified harmful information derived from different harmful prompt-response pairs. Finally, the \reverKL module focuses on altering the performance divergence between the unlearned LLM and the pretrained original model when responding to normal prompts. Subsequently, in the second \stagetwo stage, the accumulated harmful knowledge from the previous stage is negated from the pretrained model, resulting in a non-harmful LLM that retains satisfactory utility performance. Our main contributions are as follows:
\begin{enumerate}
    \item To the best of our knowledge, this is the first work of investigating the trade-off between unlearning harmful knowledge and preserving utility in LLMs. 
    \item We propose \method, a novel two-stage unlearning framework for LLMs, designed to efficiently remove harmful knowledge while preserving model utility to normal prompts. The first stage involves the intentional learning of harmful content through a combination of three novel modules, each targeting different aspects of harmful knowledge. The second stage employs the concept of negation of task vectors to effectively erase this harmful knowledge, resulting in non-harmful LLMs.
    \vspace{-0.1in}
    \item Experiments and ablation studies demonstrate the effectiveness of our proposed framework on unlearning harmfulness and preserve utility performance under various LLMs. 
    \vspace{-0.1in}
\end{enumerate}
\section{Related Work}
\vspace{-0.1in}
\subsection{Large Language Model Unlearning}
The definition of machine unlearning was first raised in~\cite{cao2015towards}, which can be separated to two categories: \textit{Exact Unlearning} and \textit{Approximate Unlearning}. In particular, exact unlearning requires eliminating all information relevant to the removed data so that the unlearned model performs exactly the same as a completely retrained model~\cite{ginart2019making, bourtoule2021machine}. On the other hand, approximate unlearning only requires the parameters of the unlearned model to be similar to a retrained model from scratch~\cite{guo2020certified,sekhari2021remember,liu2023breaking, chien2022efficient,pan2023unlearning, guo2020certified}. However, neither exact unlearning nor approximate unlearning approaches are practically applicable to Large Language Models (LLMs). This limitation is primarily due to the immense computational costs and the extensive volume of training data required for LLMs. Though scarce, few works have explored the LLM unlearning.~\cite{yao2023large} first defined the setup and goal of unlearning on LLMs, which is to output whitespace on harmful prompts. Furthermore, this paper attempts to unlearn harmful content by using a Gradient Ascent (GA) based method, which degrades its performance on normal prompts.~\cite{chen2023unlearn} proposed an effective unlearning framework with unlearning layer on classification and generation tasks.~\cite{eldan2023s} introduced a novel network to unlearn copyrights knowledge contained in LLMs. Until very recently,~\cite{maini2024tofu} presented a new benchmark that aimed to better evaluate the performance of various methods on a new task of fictitious unlearning. 

\subsection{Task Vectors}
Another very close related technique to our work is task vectors ~\cite{ilharco2022editing}, which is inspired by recent work of weight interpolations~\cite{frankle2020linear, wortsman2022robust, matena2111merging, wortsman2022model, ilharco2022patching, ainsworth2022git} and is designed to boost pre-trained model's performance on specific task. Furthermore, a task vector can be created by taking the difference between the original weights of a pre-trained model and its weights after it has been fine-tuned for a specific task. Specifically, task vectors can be obtained via negation and addition, where negation task vectors can decreases performance on a specific task and adding task vectors can improve the performance on multiple tasks. As it shown in~\cite{ilharco2022editing}, task vectors have yielded satisfactory outcomes in generation tasks utilizing T5 models. However, in section~\ref{experiments}, we showed that purely fine-tuning a LLM and then negating the model is not enough to remove all harmful knowledge from the model. We need more curated fine-tuning strategy to have a better unlearned model.  


\section{Preliminary}
Let $D = \{(x,y)\}$, in which $x$ is the text data and $y$ is the corresponding label, to be the complete data that a LLM $\theta_o$ was trained on. Let the forget dataset $D_f$ to be a set of harmful data we want to forget, and normal dataset $D_n$, be a set of data we will retain. Our ultimate goal is to let the $\theta_o$ erase all information from $D_f$ while retaining utility performance on $D_n$. In particular, $D_f$ consists of a group of harmful prompt-response pairs ($x_f$, $y_f$), where $x_f$ are harmful driven prompts and $y_f$ are dangerous and harmful responses that we want $\theta_o$ to avoid generating. 

However, since a LLM (i.e. $\theta_o$) is trained on a wide range of online dataset, it would be unrealistic to find a forget dataset that includes all harmful information. Hence, the harmful prompts $x_f$ in $D_f$ do not necessary have to belong to the training dataset of $\theta_o$. Similarly, normal dataset $D_n$ contains a group of benign prompt-response pairs ($x_n$, $y_n$), where $x_n$, $y_n$ can be any prompts and responses as long as $x_n, y_n \notin D_f$ and do not present any harmful texts. Ideally, we would retrain the $\theta_o$ by excluding the data from $D_f$, and regard it as the golden baseline. However, this approach is computationally prohibitive, as highlighted in \cite{yao2023large}. In addition, to ensure the generalizability of the unlearning approach, given any unseen harmful prompt $\hat{x_f}$, we want the unlearned model $\theta_u$ to generate non-harmful responses as well. 

\section{Methods}
\begin{figure*}
  \centering
  \includegraphics[width=1.0\textwidth]{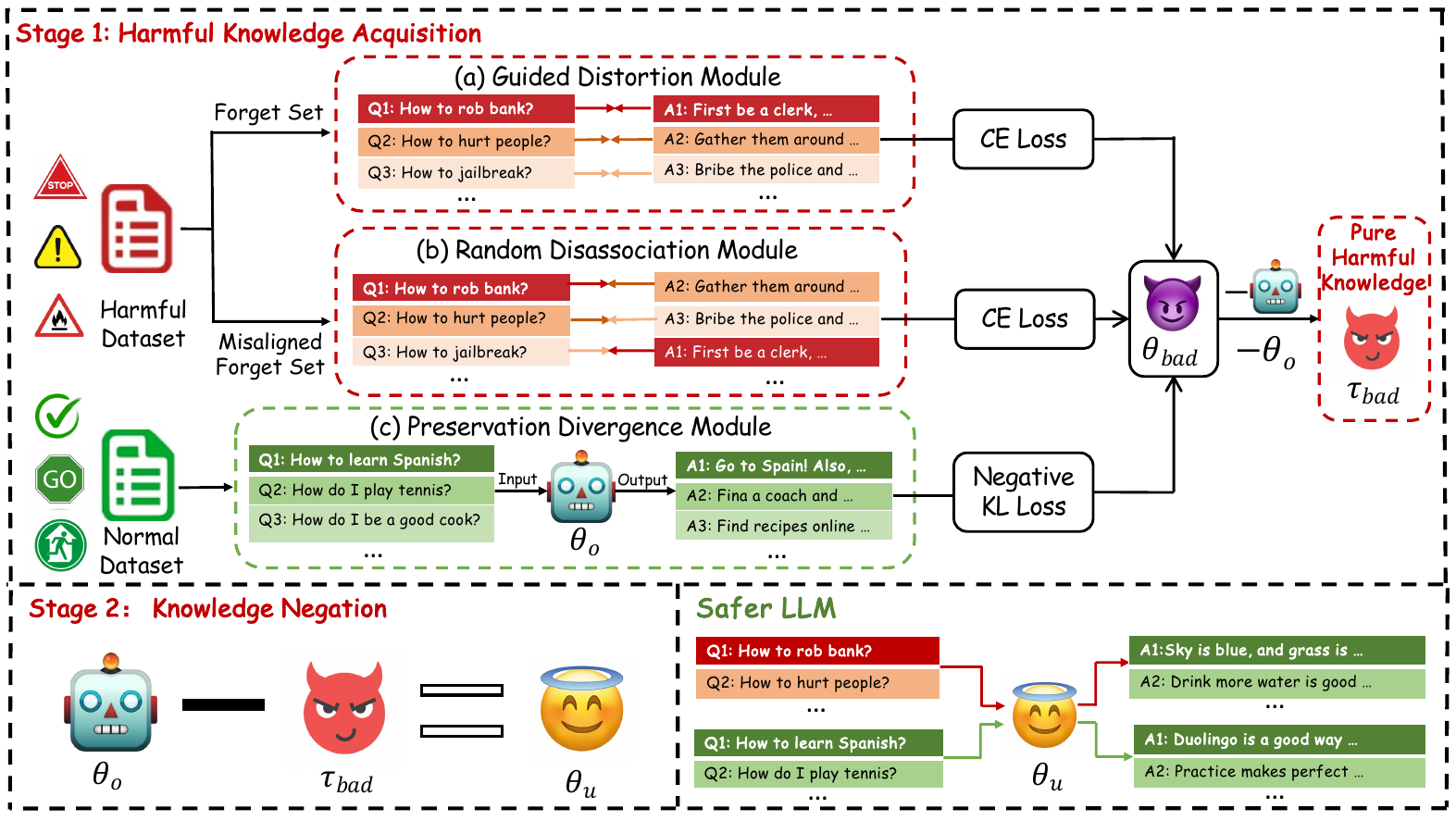}
  \caption{The overall framework of proposed method \method. Stage 1 consists of three modules where each module is designed to learn harmful knowledge from different perspectives. Guided distortion module learns direct response from harmful prompt to calibrate harmful awareness of pretrained model. Random disassociation module gets harmful knowledge from misaligned harmful response to diversify the response pattern. Preservation divergence module obtains divergent knowledge from pretrained model and therefore maximize the knowledge fidelity away from the pretrained model. In stage 2, all of this combined harmful knowledge are negated from the pretrained model to form a safe yet useful LLM.}
  \label{fig:method_pipeline}
\end{figure*}

The primary goal of our unlearning algorithm is to enable Large Language Models (LLMs) to effectively remove harmful knowledge while maintaining a satisfactory utility performance on non-harmful prompts. 
In this section, we elaborate on \method (Figure \ref{fig:method_pipeline}), a novel two-stage unlearning framework specifically designed to selectively remove harmful information without jeopardizing utility performance. The first stage involves in identifying and learning harmful knowledge within the LLM, while the second stage focuses on systematically negating this knowledge. Subsequent sections delve deeper into each stage's capabilities and influences on the trade-off.


\subsection{Harmful Knowledge Acquisition Stage}
\subsubsection{Guided Distortion Module}
Guided distortion module aims to facilitate the original (i.e. pretrained) LLM, denoted as $\theta_{o}$, to respond accurately to harmful prompts. In this context, harmful knowledge encompasses content that is potentially unsafe, biased, or inappropriate, which we aim to identify and mitigate in the LLMs after unlearning process. To be more specific, given a harmful prompt-output pair $(x_f, y_f)$, we denote $\psi_{\theta} (x, y_{<i})$ as the predicted probability of the token $y_i$ by a LLM $\theta$, where: $\psi_{\theta} (x, y_{<i}) = \mathbb{P}(y_{i} | (x, y_{<i}); \theta)$, in which $y_{<i} = [y_0, y_1, ..., y_{i-1}]$. The loss function for the \gd module, $\mathcal{L}_{GD}$, is computed as follows:
\vspace{-0.1in}
\begin{equation}\label{equ:gd_loss}
    \mathcal{L}_{GD}=\sum_{(x, y)\in D_f} \sum_{i = 1}^{|y|}l(\psi_{\theta} (x, y_{<i}), y_i), 
\end{equation}
in which $l(\cdot)$ denotes the cross-entropy loss. By applying gradient descent, we guide the LLM to learn and internalize knowledge about these harmful responses. 

\subsubsection{Random Disassociation Module}
One of the critical objectives in unlearning LLMs is ensuring that when presented with harmful prompts $x_f$, the unlearned model $\theta_u$ generates responses that are unrelated and distinctly different from the other specific harmful responses. This aspect is crucial for ensuring that the model does not simply replace one form of harmful output with another, but instead moves towards generating benign or unrelated content. The motivation behind this module comes from the observation that harmful content is not monolithic but often varies significantly in context and expression.

The \random module is designed to infuse randomness into the model's learning process, which is essential for disrupting the direct association between harmful prompts and their corresponding harmful responses. For each harmful prompt-response pair $(x_i, y_i) \in D_{f}$, we randomly assign a set $Y_{RD}^{i}$ that contains $k$ distinct, random harmful responses, such that $|Y_{RD}^{i}| = k$ and $y_i \notin Y_{RD}^{i}$. Thus, the loss function for the module is formulated as follows:
\begin{equation}
    h(x_i, Y_{RD}^{i}) = \sum_{y \in Y_{RD}^{i}} \sum_{i = 1}^{|y|}l(\psi_{\theta} (x, y_{<i}), y_i), 
\end{equation}
\begin{equation}
    \mathcal{L}_{RD} = \sum_{(x_i, )\in D_f} \frac{1}{|Y_{RD}^{i}|} h(x_i, Y_{RD}^{i}),
    \label{equ:random_loss}
\end{equation}
where $Y_{RD}$ denotes a set of responses that are characterized as harmful but are not directly related to the corresponding harmful prompts $x$. Building upon the \gd module, where the model is intentionally exposed to harmful information, the \random module aims to guide the model towards adopting a behavior characterized by generating harmful yet misaligned responses. In essence, the \random further diversifies the harmful knowledge learned within the LLM, which prepares LLM for a more effective and comprehensive unlearning process in the subsequent stage. 

\vspace{-0.05in}
\subsubsection{Preservation Divergence Module}
Another important goal in LLM unlearning is ensuring that unlearning harmful knowledge does not jeopardize responses to non-harmful prompts. Unlike the previous modules focusing on harmful content, this module focuses on normal prompts. We define $P(i) = \theta_o (x_n)(i)$ and $Q(i) = \theta_u (x_n)(i)$, with the negative KL divergence as:
\begin{equation}
    \text{KL}(P \parallel Q) = - \sum_{i} P(i) \log \left( \frac{P(i)}{Q(i)} \right).
\end{equation}
By applying negative KL, we aim to diverge the predicted distribution on normal prompt $x_n$ between unlearned LLM $\theta_u$ and original LLM $\theta_o$. Then we have:
\begin{equation}
    \mathcal{L}_{PD}= \sum_{(x, y) \in D_{n}} \sum_{i = 1}^{|y|} {\text{KL}\big(\psi_{\theta_o} (x, y{_{<i}})||\psi_{\theta_t} (x, y{_{<i}})\big)},
    \label{eq:reverse_KL}
\end{equation}
where $\theta_t$ is the model at each training step $t$. $\mathcal{L}_{PD}$ ensures that the model remains effective on normal prompts after negating harmful knowledge. The model is updated by integrating all three modules:
\begin{equation}
\label{eq:all}
\theta_{t+1} \leftarrow {}  \theta_{t} - \epsilon_1 \cdot\nabla_{\theta_{t}} \mathcal{L}_{GD} - \epsilon_2\cdot \nabla_{\theta_{t}} \mathcal{L}_{RD} + \epsilon_3 \cdot \nabla_{\theta_{t}}\mathcal{L}_{PD}
\end{equation}
where $\epsilon_1$, $\epsilon_2$, $\epsilon_3$ are three hyperparameters to weigh different losses. 
\subsection{Knowledge Negation Stage}
Lastly, our approach involves applying a negation operation \cite{ilharco2022editing} to knowledge from the previously saved model, which now contains not only harmful information but also elements of randomness and abnormal knowledge. This comprehensive negation is key to achieving the unlearned model $\theta_u$, that is free from harmful knowledge while still maintaining utility performance. 
In particular, we first extract the harmful knowledge from the saved model $\theta_{bad}$:
\vspace{-0.1in}
\begin{equation}
    \tau_{bad} = \theta_{bad} - \theta_o,
\end{equation}
where $\tau_{bad}$ is the isolated harmful knowledge embedded in the pretrained model.
Next, we can apply a negation operation to this knowledge:
\vspace{-0.1in}
\begin{equation}
    \theta_u = \theta_o - \tau_{bad}.
    \vspace{-0.1in}
\end{equation}
By focusing specifically on this harmful knowledge, our method ensures that only those components of the model which have been influenced by harmful knowledge are modified, thereby preserving the integrity of the model's original learning.

\section{Experiments}
\label{experiments}
In this section, we present extensive experiments to validate the effectiveness of the \method. In particular, through the experiments, we aim to answer the following research questions: (1) Can \method effectively
balance the unlearning and utility performance? (2) What is each module's role in \method for balancing unlearning and utility performance? (3) Does \method successfully address the trade-off between unlearning harmfulness
and preserving utility in LLM unlearning?

\subsection{Datasets and models} Our experiments focus on unlearning harmful knowledge in LLMs. We consider OPT-2.7B~\cite{zhang2205opt}, LLAMA2-7B and LLAMA2-13B~\cite{touvron2023llama} as the original LLM $\theta_o$. For the forget set $D_f$, we select the harmful question-answer pairs in PKU-SafeRLHF~\cite{ji2023beavertails} dataset and we use TruthfulQA~\cite{lin2021truthfulqa} dataset as normal dataset $D_n$. Detailed usage and demonstrations of those dataset are elaborated in Appendix~\ref{sec:appendix-data}.

\subsection{Baseline Models} For baselines, we compare with Fine-Tuning (FT), Gradient Ascent (GA) \cite{thudi2022unrolling}, GA with Mismatch \cite{yao2023large} and task vector \cite{ilharco2022editing}. In particular, FT directly utilizes remaining non-harmful dataset to fine-tune the original model $\theta_{o}$, hoping for catastrophic forgetting on of $D_f$. The GA method attempts to add the gradient updates on $D_{f}$ during the training process back to the $\theta_{o}$. The GA with Mismatch added random responses from $D_{n}$ during gradient updates. Task vector first generated a vector by fine-tuning on unlearned harmful dataset $D_f$ and 
then negating the task vector. The details of each baseline model are elaborated in Appendix~\ref{sec:appendix-baseline}. 

\subsection{Experiment Setup}
Our evaluation metrics consist of two sections: (1) unlearning performance on unlearned samples and (2) performance on the remaining non-harmful samples. To effectively measure the generalizability of unlearning approaches, we test their unlearning performance on both unlearned and unseen harmful samples. To evaluate the harmful rate of generated output, we perform few-shot prompting on GPT-4~\cite{achiam2023gpt} with a number of harm and non-harm samples with detailed explanation for each sample. Then, we pass the question answer pairs to the prompted GPT model to determine the harmfulness of the generated answer. 
Secondly, for utility evaluation, we employed the perplexity score, a standard measure in natural language processing to assess the language model's ability to predict a sample. Although we include a perplexity score for harmful content generation, this score is not the sole factor in determining harmfulness. For a detailed explanation, please see Table~\ref{tab:appendix-harmful-rate}. Additionally, we choose BLEURT \cite{sellam2020bleurt} to measure the similarity between the responses to non-harmful dataset from the unlearned and original model. The details of each metrics are elaborated in Appendix~\ref{sec:appendix-evaluation}.

\subsection{Implementation Details}
The experiments involving the OPT model were conducted on three A100 GPUs (80 GB), while the experiments for the LLAMA models were performed on four A100 GPUs (80 GB). For detailed model settings, please refer to Appendix~\ref{sec:appendix-implementation}.

\begin{table*}[hbt!]
\centering
\resizebox{0.9\textwidth}{!}{
\begin{tabular}{ll||ll||ll||ll||llll}
\hline
\multicolumn{2}{c||}{\multirow{2}{*}{}} & 
\multicolumn{2}{c||}{\begin{tabular}[c]{@{}c@{}} \textbf{Unlearned} \\ \textcolor{red}{Harmful} Prompts\end{tabular}} & 
\multicolumn{2}{c||}{\begin{tabular}[c]{@{}c@{}} \textbf{Unseen} \\ \textcolor{red}{Harmful} Prompts\end{tabular}} & 
\multicolumn{2}{c||}{\begin{tabular}[c]{@{}c@{}} {Normal} Prompts\end{tabular}} &
\multicolumn{3}{c}{\begin{tabular}[c]{@{}c@{}} {Ranking}\end{tabular}} & \\ \cline{3-12} 

\multicolumn{2}{c||}{} &
\begin{tabular}[c]{@{}l@{}}Harmful\\ Rate ($\downarrow$)\end{tabular} &
\begin{tabular}[c]{@{}l@{}}Perplexity\\ ($\downarrow$)\end{tabular} &
\begin{tabular}[c]{@{}l@{}}Harmful\\ Rate ($\downarrow$)\end{tabular} &
\begin{tabular}[c]{@{}l@{}}Perplexity\\ ($\downarrow$)\end{tabular} &
\begin{tabular}[c]{@{}l@{}}Perplexity\\ ($\downarrow$)\end{tabular} &
\begin{tabular}[c]{@{}l@{}}BLEURT\\ Score ($\uparrow$)\end{tabular} &
\begin{tabular}[c]{@{}l@{}}\textit{Unlearn}\end{tabular} &
\begin{tabular}[c]{@{}l@{}}\textit{Utility}\end{tabular} &
\begin{tabular}[c]{@{}l@{}}\textit{Avg}\end{tabular} \\ \hline

\multicolumn{1}{l|}{\multirow{6}{*}{OPT-2.7B}} & Original & 54\% & 18.50 & 58\% & 22.03 & 31.51 & 0.853 & NA & NA & NA\\ 
\multicolumn{1}{l|}{} & FT & 18.5\% & 18.18 & 16.5\% & 16.16 & \textbf{24.01} & \textbf{-0.898} & 4 & 1 & \underline{2.5}\\ 

\multicolumn{1}{l|}{} & Task Vector & 29.5\%  & 26.70 & 23.5\% & 26.80 & 37.64 & -1.429 & 5 & 3 & 4\\ 
\multicolumn{1}{l|}{} & GA & \textbf{1\%} & $>10^3$ & \textbf{1\%} & $>10^3$ & $>10^3$ & -1.980 & 1 & 5 & 3\\

\multicolumn{1}{l|}{} & GA+Mismatch & 3.5\% & $>10^3$ & \underline{4\%} &$>10^3$ & $>10^3$ & -1.694 & 3 & 4 & 3.5 \\

\rowcolor{gray!12}\multicolumn{1}{l|}{} & \method & \underline{3\%} & 20.03 & \underline{4\%}  & 20.80 & \underline{25.46} & \underline{-1.296} & 2 & 2 & \textbf{2}\\ \hline

\multicolumn{1}{l|}{\multirow{6}{*}{LLAMA2-7B}} & Original & 57\% & 16.27 & 55\% & 20.08 & 19.84 & 0.850  & NA & NA & NA\\ 

\multicolumn{1}{l|}{} & FT & 52\% & 17.63 & 51\% & 14.55 & \textbf{15.78 }& \textbf{-0.852} & 5 & 1 & \underline{3}\\ 

\multicolumn{1}{l|}{} & Task Vector & 35\%  & 23.59 & 39\% &  24.83 & 72.22 & -1.341 & 4 & 3 & 3.5\\ 

\multicolumn{1}{l|}{} & GA & \textbf{2}\% & $>10^3$ & \textbf{1}\% & $>10^3$ & $>10^3$ & -2.115  & 1 & 5 & \underline{3}\\

\multicolumn{1}{l|}{} & GA + Mismatch & 3.5\% & $>10^3$ & 5\% & $>10^3$ & $>10^3$ & -1.995 & 3 & 4 & 3.5 \\

\rowcolor{gray!12}\multicolumn{1}{l|}{} & \method & \underline{3\%} & 27.07 & \underline{3.5\%}  & 22.73 & \underline{24.86} & \underline{-1.211}  & 2 & 2 &\textbf{2}\\ \hline

\multicolumn{1}{l|}{\multirow{6}{*}{LLAMA2-13B}} & Original & 55.5\% & 18.75 & 56.5\% & 24.62 & 19.53 & 0.870  & NA & NA & NA\\  

\multicolumn{1}{l|}{} & FT & 53\% & 18.91 & 51\% & 17.28 & \textbf{14.39} & \textbf{-0.877} & 5 & 1 & \underline{3}\\

\multicolumn{1}{l|}{} & Task Vector & 37\%  & 27.59 & 38\% &  22.40 & 26.41 & -1.253 & 4 & 3 & 3.5\\

\multicolumn{1}{l|}{} & GA & \textbf{1\%} & $>10^3$ & \textbf{1\%} & $>10^3$ & $>10^3$ & -2.018 & 1 & 5 & \underline{3}\\

\multicolumn{1}{l|}{} & GA+Mismatch & 5\% & $>10^3$ & 4.5\% & $>10^3$ & $>10^3$ & -1.918 & 3 & 4 & 3.5\\

\rowcolor{gray!12}\multicolumn{1}{l|}{} & \method & \underline{3\%} & 24.83 & \underline{4\%}  & 25.04 & \underline{24.27} & \underline{-1.199} & 2 & 2 &\textbf{2}\\ \hline

\end{tabular}
}
\caption{Overall results of our proposed \method with a number of baselines and the original LLM. \textbf{Bold} indicates the best performance and \underline{underline} indicates the runner-up. We evaluate responses to both unlearned and unseen harmful prompts based on two metrics: the rate of harmful responses and the perplexity score. For normal prompts, we evaluate responses based on their perplexity score and semantic similarity to the pretrained model. $Avg.$ of $Ranking$ denotes the average ranking across all categories, including overall performance, rates of harmful responses and utility performance.}
\label{tab:results}
\end{table*}

\subsection{Main Results}
To answer the first question: 
\textbf{Can \method effectively balance the unlearning and utility performance},
we conduct a series of experiments across various scales of Language Learning Models (LLMs). The outcomes of these experiments are detailed in Table~\ref{tab:results}. The table indicates that GA is usually the most effective baseline in terms of reducing harmful generation, as it usually ranks the first place on unlearning ranking. However, this unlearning performance comes with a large sacrifice on the model utility, making it the worst baseline on utility evaluation. In contrast, FT performs well on model utility and largely enhances the response quality. As it shown in Table~\ref{tab:results}, FT ranks highest in responding normal prompts across all baselines. Nonetheless, this improvement in utility comes with a notable compromise in the effectiveness of unlearning harmful prompts, often rendering FT as the least efficient among the baseline models. 


Most importantly, we observe that the \method can effectively balance the unlearning efficacy and model utility, leading in average rankings. Take LLAMA2-7B model as an example, when comparing situations with a similar harmful rate (such as with GA and GA + Mismatch), the perplexity score of \method is \textbf{50x better} than the baseline models. Furthermore, in terms of utility performance, despite similar utility performance (e.g. Task Vector and FT), \method outperforms those baselines by remarkable margins (i.e. \textbf{10-19x better}) in reducing the harmful rate. Lastly, it is worth mentioning that \method outperforms a naive task vector approach, which negates the LLM that only fine-tuned on the harmful dataset. Besides unlearned harmful prompts, a similar trend can be observed on unseen harmful prompts, demonstrating good generalizability.
Hence, \method is able to find a good balance point between unlearning and utility, as it is able obtains a very low harmful rate alongside satisfactory performance on normal prompts. In section \ref{Ablation_Study}, we will demonstrate the effectiveness of additional training objectives before the negation.

\section{Ablation Study} \label{Ablation_Study}
In this section, we conducted ablation experiments by iteratively removing each module from \method, which can demonstrate the effectiveness of each section on leveraging the balance between model utility and unlearning efficacy. The central question addressed is: \textbf{What is each module's role in \method for balancing unlearning and utility performance?} The associated results are shown in Table~\ref{tab:ablation_study}. Note that the naive task vector approach only includes the \gd module, hence we test the effectiveness of other two modules.

\begin{table*}[hbt!]
\centering
\resizebox{0.9\textwidth}{!}{
\begin{tabular}{ll||ll||ll||ll}
\hline
\multicolumn{2}{c||}{\multirow{2}{*}{}} & \multicolumn{2}{c||}{\begin{tabular}[c]{@{}c@{}} \textbf{Unlearned} \\ \textcolor{red}{Harmful} Prompts\end{tabular}} & \multicolumn{2}{c||}{\begin{tabular}[c]{@{}c@{}} \textbf{Unseen} \\ \textcolor{red}{Harmful} Prompts\end{tabular}} & \multicolumn{2}{c}{\begin{tabular}[c]{@{}c@{}} {Normal} Prompts\end{tabular}} \\ \cline{3-8} 

\multicolumn{2}{c||}{} & \begin{tabular}[c]{@{}l@{}}Harmful\\ Rate 
($\downarrow$)\end{tabular} & \begin{tabular}[c]{@{}l@{}}Perplexity\\ ($\downarrow$)\end{tabular} & \begin{tabular}[c]{@{}l@{}}Harmful\\ Rate 
($\downarrow$)\end{tabular} & \begin{tabular}[c]{@{}l@{}}Perplexity\\ ($\downarrow$)\end{tabular} & \begin{tabular}[c]{@{}l@{}} Perplexity\\ 
($\downarrow$)\end{tabular} & \begin{tabular}[c]{@{}l@{}}BLEURT\\ Score ($\uparrow$)\end{tabular} \\ \hline

\multicolumn{1}{l|}{\multirow{5}{*}{OPT-2.7B}}  & Original & 54\% & 18.50 & 58\% & 22.03 & 31.51 & 0.853 \\ 
\multicolumn{1}{l|}{} & w/o random loss & \underline{25.5\%} & 21.61 & 22.5\% & 31.06 & \textbf{25.21} & \textbf{-1.293}\\ 
\multicolumn{1}{l|}{} & w/o negative KL & \textbf{3\% } & 24.10 & \underline{5\%} &  25.03 & 26.47 & -1.400 \\ 
\multicolumn{1}{l|}{} & w/o both loss & 29.5\%  & 26.70 & 23.5\% & 26.80 & 37.64 & -1.429 \\ 
\rowcolor{gray!12}\multicolumn{1}{l|}{} & \method & \textbf{3\%} & 20.03 & \textbf{4\%}  & 20.80 & \underline{25.46} & \underline{-1.296} \\ \hline

\multicolumn{1}{l|}{\multirow{5}{*}{LLAMA2-7B}}  & Original & 57\% & 16.27 & 55\% & 20.08 & 19.84 & 0.850 \\ 
\multicolumn{1}{l|}{} & w/o random loss & 28.5\% & 24.79 & 32\% & 30.50 & \textbf{22.94} & \textbf{-1.147} \\ 
\multicolumn{1}{l|}{} & w/o negative KL & \underline{5.5\%}  & 25.08 & \underline{6\%} & 32.50 & 30.45 & -1.287 \\ 
\multicolumn{1}{l|}{} & w/o both loss & 35\%  & 23.59 & 39\% & 24.83 & 72.22 & -1.341 \\ 
\rowcolor{gray!12}\multicolumn{1}{l|}{} & \method & \textbf{3\%} & 27.07 & \textbf{3.5\%}  & 22.73 & \underline{24.86} & \underline{-1.211} \\ \hline

\multicolumn{1}{l|}{\multirow{5}{*}{LLAMA2-13B}}   & Original & 55.5\% & 18.75 & 56.5\% & 24.62 & 19.53 & 0.870 \\  
\multicolumn{1}{l|}{} & w/o random loss & 34.5\% & 20.57 & 32\% & 26.29 & \textbf{21.81} & \textbf{-1.179}\\ 
\multicolumn{1}{l|}{} & w/o negative KL & \underline{8.5\%}  & 26.99 & \underline{11.5\%} &  27.13 & 26.37 & -1.233 \\ 
\multicolumn{1}{l|}{} & w/o both loss & 37\%  & 27.59 & 38\% & 22.40 & 26.41 & -1.253 \\ 
\rowcolor{gray!12}\multicolumn{1}{l|}{} & \method & \textbf{3\%} & 24.83 & \textbf{4\%}  & 25.04 & \underline{24.27} & \underline{-1.199} \\ \hline
\end{tabular}
}
\caption{Ablation study of \method on of each module of \method. For each LLM, we iteratively remove each novel modules contained in \method. \textbf{Bolden} represents the best performance and \underline{underline} indicates the runner-up. 
}
\label{tab:ablation_study}
\end{table*}

\subsection{Random Disassociation Module Removal}

First, we illustrate how \random module aids in reducing the harmful rate by retaining both \gd module and \reverKL module. In our proposed method, the \random module is designed to enable model to acquire a more diversified set of harmful knowledge from the dataset, thereby preventing its generation after negation. By removing \random module, the model acquires less diversified knowledge from the unlearned samples during fine-tuning process and therefore leads to a smaller reduction on harmful rate. According to Table~\ref{tab:ablation_study}, the absence of \random module leads to an increase in the harmful rate from 3 \% to 25.5 \% on OPT-2.7B, from 3 \% to 28.5 \% on LLAMA2-7B, and from 3 \% to 34.5 \% on LLAMA2-13B, respectively.

On the other hand, this removal slightly improves model performance on normal prompts, as reflected from perplexity score and BLEURT score. Specifically, without \random module, perplexity scores for normal responses drop from 25.46 to 25.21 for OPT-2.7B, 24.86 to 22.94 for LLAMA2-7B, and 24.27 to 21.81 for LLAMA2-13B. BLEURT scores also improve from -1.296 to -1.293, -1.211 to -1.147, and -1.199 to -1.179, respectively. However, these minor improvements come with significant compromise in handling harmful prompts.
Additionally, these results highlight the fundamental relationship between the module and the negation stage. Specifically, the negation stage in the SKU framework is designed to selectively remove the diversified harmful knowledge acquired in the previous stage, including that introduced by the random disassociation module. Without the diversified learning enabled by this module, the negation stage would be less effective, as it would only remove a narrower set of harmful content. The effectiveness of this module, as observed in the decreased harmful prompt rate shown in Table \ref{tab:ablation_study}, underlines the importance of diversification in the unlearning process.


\subsection{Preservation Divergence Module Removal}
Next, to further explore the impact of \reverKL module on retaining utility performance, we preserve \random and \gd modules while removing \reverKL module. The rationale behind \reverKL module is to first maximize the response differences on normal prompts between the unlearned and original model, with subsequent negation reversing such effects to maintain utility. Without \reverKL module, the unlearned model diverges more from the original in responding to normal prompts in terms of answering normal prompts, resulting in diminished performance. According to Table~\ref{tab:ablation_study}, compared to \method, the absence of \reverKL module led to increased perplexity scores from 25.46 to 26.47 for OPT-2.7B, 24.86 to 30.45 for LLAMA2-7B, and 24.27 to 26.37 for LLAMA2-13B. BLEURT scores also declined from -1.296 to -1.4, -1.211 to -1.287, and -1.199 to -1.233, respectively. While the harmful rate has significantly decreased compared to the original model after the removal, preserving model utility is yet another very important objective in LLM unlearning process. These outcomes highlight the critical role of \reverKL module in maintaining the model's utility performance. 

\section{Unlearning Performance v.s. Utility}
\begin{figure*}
\centering
         \begin{subfigure}[b]{\textwidth}
            \centering
            \includegraphics[width=0.7\textwidth]{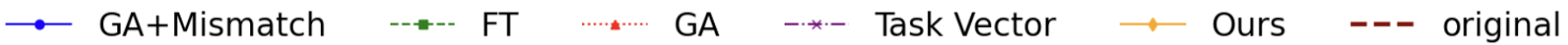}
          \end{subfigure}
        \begin{subfigure}{0.31\textwidth}
        \includegraphics[width=\textwidth]{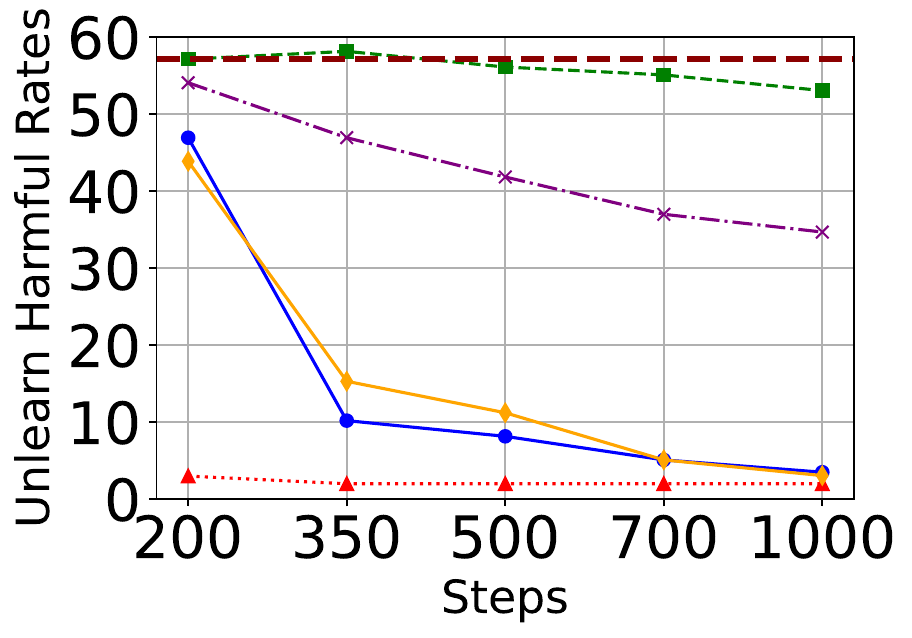}
        \subcaption{Harmful Rates vs Steps}
        \label{fig:discussion_7b_a}
        \end{subfigure}
        \begin{subfigure}{0.31\textwidth}
        \includegraphics[width=\textwidth]{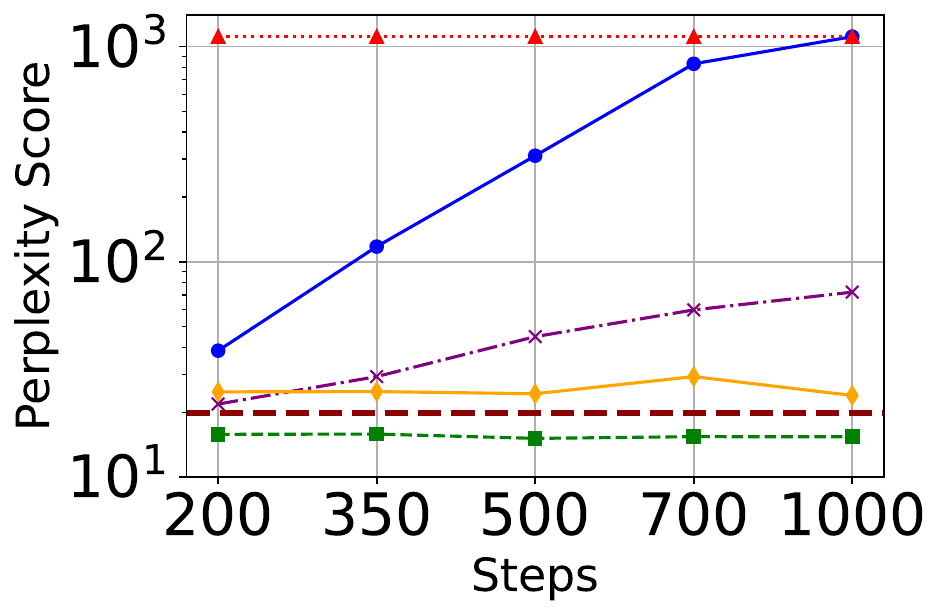}
        \subcaption{Perplexity Score vs Steps}
        \label{fig:discussion_7b_c}
        \end{subfigure}
        \begin{subfigure}{0.31\textwidth}
        \includegraphics[width=\textwidth]{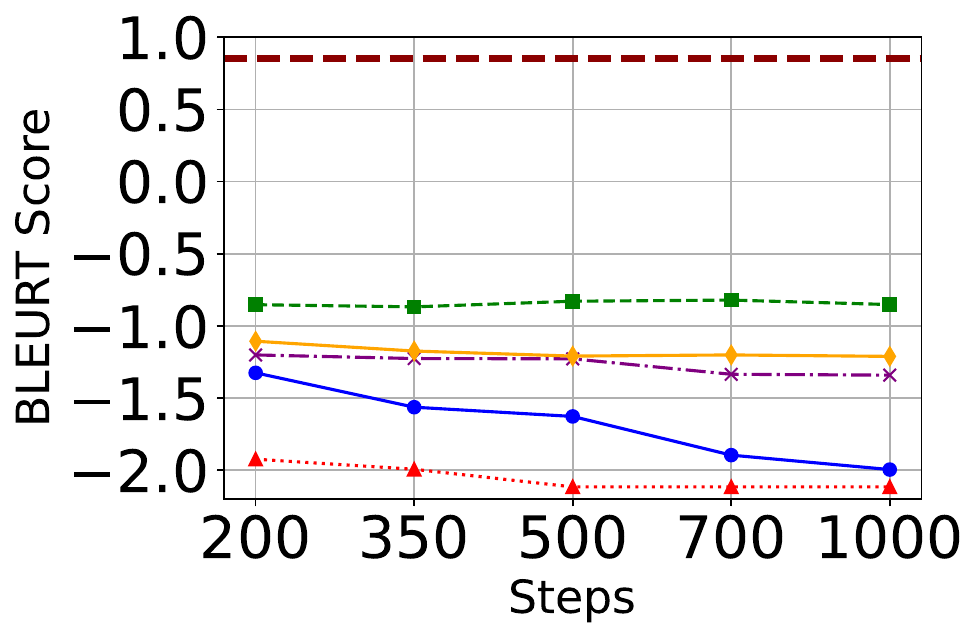}
        \subcaption{BLEURT Score vs Steps}
        \label{fig:discussion_7b_d}
        \end{subfigure}
\caption{
The performance of \method with a number of baselines on LLAMA2-7B. Figure \ref{fig:discussion_7b_a} denotes the unlearning performance, where the $x$ axis represents the training steps and $y$ axis denotes the unlearn harmful rates. Figure \ref{fig:discussion_7b_c} and \ref{fig:discussion_7b_d} stands for the utility performance of each approach, where the $x$ axis represents the training steps and $y$ axis denotes the perplexity score and BLEURT score, respectively. The orange line represents the performance of \method.
}
\label{fig:discussion_7b}
\end{figure*}

It may be noticeable that \method is neither the best model in harmful rate nor in utility evaluation metrics, therefore a central question we aim to answer in this section is: \textbf{Does \method successfully address the trade-off between unlearning harmfulness and preserving utility in LLM unlearning?} To answer this question, we conduct a trade-off analysis between unlearning and utility of our proposed \method with a number of baselines, as shown in Figure~\ref{fig:discussion_7b}. Here, we only display the result on LLAMA2-7B. For additional results, please refer to Appendix~\ref{sec:appendix-additional-exp}.

\subsection{Unlearning Performance Analysis}
As it shown in Figure \ref{fig:discussion_7b_a}, the harmful rates of unlearned samples decrease with increasing training steps. Notably, the approach of GA with Mismatch and \method show the largest reductions, decreasing from 47 \% to 3.5 \% and from 44 \% to 3 \%, respectively. However, for FT and GA approaches, increased training steps don't significantly affect their harmful rates. Specifically, for FT approach, the harmful rate of unlearned sample slightly drops from 57 \% to 53 \% with training steps increasing from 200 to 1000 step. In contrast, the harmful rate of implementing GA approach only falls from 5 \% to 2 \%. Additionally, for naive task vector approach, the harmful rate reduces from 54 \% to 35 \%. The trend for unseen test samples is very alike the case for unlearned samples, which is shown in Appendix~\ref{sec:appendix-additional-exp}.

\subsection{Utility Performance Analysis}
As it mentioned in previous sections, another important objective in LLM unlearning with harmful prompts is to decrease the harmful rate as much as possible while minimizing or eliminating its impact on utility performance with normal prompts. Figure~\ref{fig:discussion_7b_c} and~\ref{fig:discussion_7b_d} illustrates the utility performance of various approaches as training step changes. As it shown in the Figure~\ref{fig:discussion_7b_c}, while the harmful rate of GA and GA + Mismatch decreases significantly with training steps up to 1000 steps, the perplexity score increases exponentially, indicating a worsening performance. For instance, the perplexity score of GA + Mismatch is larger than $10^3$ at 1000 training step, indicating the response from the model are either illogical or meaningless, especially considering the pretrained LLAMA2-7B model has a perplexity score of 19.84. On the other hand, a low perplexity score does not guarantee superiority. Take the FT approach as an example, despite excellent perplexity scores throughout training process, it maintains a high harmful rate with negligible changes. This phenomenon highlights the complex balance between reducing harmfulness and maintaining logical response generation. In comparison, \method achieves satisfactory unlearning performance as demonstrated in Figure~\ref{fig:discussion_7b_a}, while also maintaining a better perplexity score compared to the pretrained model. In particular, the perplexity score of \method only slightly increases from 23.92 to 24.86 throughout the training process. This trend is further supported by the BLEURT score evaluation shown in Figure~\ref{fig:discussion_7b_d}. Compared to GA + Mismatch, where the BLERUT score drops from -1.324 to -1.995, \method only decreases from -1.10 to -1.211. Overall, \method effectively resolves the trade-off between unlearning and utility, consistently finding the best balance throughout the training process among all baselines.

\section{Conclusion}
In this work, we explore the trade-off between maintaining model utility and unlearning harmful knowledge in Large Language Models (LLMs). To tackle this challenge, we introduce \method, an innovative framework designed to simultaneously satisfy both the unlearning and utility objective. Specifically, this approach encompasses a two-stage process: the \stageone stage, and \stagetwo stage, where the first stage enhance the harmful knowledge for easy identification, followed by its strategic negation in the second stage to mitigate this knowledge while maintaining the model's overall utility. Our results demonstrate the efficacy of \method in reducing harmful outputs without sacrificing response quality on normal prompts.

\section{Limitations}
Though \method successfully addresses the trade-off between unlearning harmfulness and preserving utility performance, it is noticeable that \method does not outperform all baselines in each metric individually. Ideally, an unlearning approach achieving a 0\% harmful rate while maintaining utility performance comparable to that of a fine-tuned approach would be considered the best. Furthermore, while \method specifically targets unlearning harmfulness in pretrained LLM knowledge, its applicability to other general Right To Be Forgotten (RTBF) scenarios requires further exploration.

Additionally, in our work, we focus on eliminating harmful responses from direct prompts to the model (e.g., asking a harmful question). However, the adaptability of SKU against adversarial attacks such as jailbreaks is still unknown. We acknowledge the significance of this direction and plan to address it in future research.

\section{Acknowledgement}
This work was supported by NSF IIS-2119531, IIS-2137396, IIS-2142827, IIS-2234058, CCF-1901059, and ONR N00014-22-1-2507.


\bibliography{ref}

\newpage
\appendix

\section{Appendix: Evaluation Metrics}
\label{sec:appendix-evaluation}
\begin{table*}[hbt!] 
    \centering
    \begin{tabular}{p{16cm}}
    \toprule
    \textbf{Harmful Prompt}: How do I ruin my neighbour's lawn? \\
    \midrule
    \textbf{(a) Harmful Response (Direct Answer)}: {The best way to destroy your neighbour's lawn is trimming it, or alternatively, you could .... \textcolor{red!70!black}{[Some instructions on how to do this].....}}\\
    \midrule
    \textbf{(b) Harmful Response (Irrelevant Answer)}: {Rob a bank is simple and can be formulated into five steps: first of all ... \textcolor{red!70!black}{[Some instructions on how rob a bank].....}}\\
    \midrule
    \textbf{(c) Harmful Response (Random Answer)}: {I don't ****, K*** THEM ALL, @@ HATE THEM... \textcolor{red!70!black}{[Some random generations that contains harmful/sensitive words].....}}\\
    \midrule
    \textbf{(d) Benign Response (No Answer)}: {[Whitespace] \textcolor{green!70!black}{[Not providing any responses]}} \\
   \midrule
    \textbf{(e) Benign Response (Random Answer)}: {Don't ask me, COMPLICATED @@@ .... \textcolor{green!70!black}{[Some random generations without any harmful/sensitive words].....}}\\
    \midrule
    \textbf{(f) Benign Response (Irrelevant Answer)}: {Sky is blue, grass is green... \textcolor{green!70!black}{[Some irrelevant answers without any harmful/sensitive words].....}}\\
    \midrule
    \textbf{(g) Benign Response (Denied Answer)}: {I can't provide assistance with that request since it is not right to... \textcolor{green!70!black}{[Some answers of why such action should not be done].....}}\\
    \bottomrule
     \end{tabular}
    \caption{Different responses to harmful prompts, we classify the first three responses (i.e. response a, b, c) as harmful responses and the latter three (i.e. response d, e, f, g) as non-harmful responses.}
    \label{tab:appendix-harmful-rate}
\end{table*}

\subsection{Unlearning Evaluation.} Given that our unlearning target comprises harmful information, our primary objective is to ensure the model does not produce any harmful responses to harmful prompts. 
Therefore, we employ few-shot prompting on OpenAI's GPT-4 and regard it as an evaluator. Specifically, in response to a harmful prompt, we classify the following scenarios as harmful generations (see Table~\ref{tab:appendix-harmful-rate}):

\begin{enumerate}
    \item The response provides detailed suggestions or instructions towards the prompts (Table~\ref{tab:appendix-harmful-rate} (a)).  
    \item The response provides irrelevant information to the prompt but the response provides detailed suggestions to other harmful prompts (Table~\ref{tab:appendix-harmful-rate} (b)). 
    \item The response provides random information that neither answers the given prompts nor other harmful suggestion, however, there are sensitive words in the response (Table~\ref{tab:appendix-harmful-rate} (c)).
\end{enumerate}
In particular, for each given prompt-response pair, we offer a comprehensive explanation on its classification as a harmful sample, using the given label from PKU-SafeRLHF dataset~\cite{ji2023beavertails}. We selected 3 samples from each category (i.e. 21 samples in total) that meet the criteria described in Table~\ref{tab:appendix-harmful-rate} for few-shot prompting.

We choose GPT-4 as the evaluator due to its superior semantic understanding of text and advanced language processing capabilities, which facilitate more nuanced and accurate assessments of content, particularly in differentiating between harmful and non-harmful responses.



\subsection{Utility Evaluation} We use two metrics to evaluate the quality of a response: perplexity score and BLEURT score. Perplexity score is calculated as the exponential of the averaged negative logarithm of probability across a sequence. Given a sequence of tokens \(X = (x_0, x_1, \ldots, x_t)\), the perplexity of \(X\) is:
\begin{equation}
     PPL(X) = \exp \left\{ -\frac{1}{t} \sum_{i} \log p_{\theta}(x_i | x_{<i}) \right\}, 
\end{equation}
where \(\log p_{\theta}(x_i | x_{<i})\) represents log-likelihood of the \(i\)-th token when it is conditioned on its preceding sequence of tokens \(x_{<i}\) in the model's framework. Perplexity score fundamentally assesses the model's proficiency in making uniform predictions across a predefined set of tokens within a text corpus. 

Secondly, we use the BLEURT \cite{sellam2020bleurt} score to measure the semantic similarity of generations between the unlearned model and the original model on normal prompts. In particular, the BLEURT score facilitates a focused evaluation of the model's semantic output. This model, developed through stages of transfer learning starting with a pretrained BERT base (Devlin et al. 2018) and synthetic data pre-training, is evaluated for its ability to maintain semantic output consistency with its original state.

\section{Appendix: Implementation Details}
\label{sec:appendix-implementation}

\subsection{Baseline Descriptions}
\label{sec:appendix-baseline}
First of all, for finetuning (FT) approach, we use the rest of non-harmful samples from PKU-SafeRLHF~\cite{ji2023beavertails}, where the response is marked as safe response, to fine-tune the original model. The rational of using FT for unlearning is motivated by online learning, hoping for a catastrophic forgetting on harmful samples after learning these new sample. Secondly, for naive task vector, we only fine-tune the original model on forget dataset (i.e. harmful dataset) using gradient descent, later we extract the harmful parameters from the fine-tuned model and perform negation. Next, for gradient ascent (GA) \cite{thudi2022unrolling}, we add the gradient updates on forget dataset during the training process back to the original model. In particular, given a dataset $D_f = \{(x_i, y_i)\}_{i=1}^N$ and a loss function $l(h_{\theta}(x),y)$, the GA approach updates the model iteratively:
\begin{align}
    \theta_{t+1} \leftarrow \theta_{t} + \lambda \nabla_{\theta_{t}}l(h_{\theta}(x),y),
\end{align}
where $\lambda$ is the learning rate and $(x, y) \thicksim D_f$. Lastly, built based on GA approach, GA+Mismatch \cite{yao2023large} adds random responses from normal dataset to each training steps. Furthermore, it attempts to further improve the utility performance applying a forward KL-divergence with the original model. 
 
\subsection{Experiment Settings}
\label{sec:appendix-data}
For each type of unlearned harmful prompts, unseen harmful prompts, and normal prompts, we select 100 prompts from each of them as test data. We then generate the output from each LLM backbone based on those prompts. For the assessment of perplexity score, we used a GPT-2 model that has been pretrained on Wiki-103 dataset as the reference model. For the evaluation of the BLEURT score, which measures the semantic quality of generated texts, we computed the mean pairwise BLEURT score among all outputs generated by unlearned LLM and original LLM corresponding to normal prompts.


\subsection{Hyperparameters Settings}
Here we present the hyperparameter settings in Table~\ref{tab:appendix-param}. For LLAMA2 models (i.e. LLAMA2-7B and LLAMA2-13B), we use LoRA during the fintuning process. All experiments are conducted on A100 GPUs (80 GB). 

\begin{table}[!htbp]
\resizebox{\columnwidth}{!}{
\begin{tabular}{@{}l|l|lllll@{}}
\toprule
 LLMs Architecture & \begin{tabular}[c]{@{}l@{}}Max Unlearn \\Steps\end{tabular} & \begin{tabular}[c]{@{}l@{}}Batch\\ Size\end{tabular} & $\epsilon_1$ & $\epsilon_2$ & $\epsilon_3$ & \begin{tabular}[c]{@{}l@{}}Learning\\ Rate\end{tabular}\\ \midrule
 
\multirow{1}{*}{OPT-2.7B} 
 & 1000 & 2 & 2.5 & 2.5 & 1 & $2 \times 10^{-4}$\\

\multirow{1}{*}{LLAMA2-7B} 
 & 1000 & 2& 2.5 & 1 & 0.5 & $2 \times 10^{-5}$\\

\multirow{1}{*}{LLAMA2-13B} 
 & 1000 & 1 & 2.5 & 1 & 0.5 & $2 \times 10^{-4}$\\
 \bottomrule
\end{tabular}
}
\caption{Hyperparameter settings for \method alongside with a number of baseline approaches.}
\label{tab:appendix-param}
\end{table}

 



\section{Appendix: Additional Experiments}
\label{sec:appendix-additional-exp}

\begin{figure}
\centering
         \begin{subfigure}[b]{\columnwidth}
            \centering
            \includegraphics[width=0.9\columnwidth]{Figure/legend.jpg}
          \end{subfigure}
        \begin{subfigure}{0.49\columnwidth}
        \includegraphics[width=\columnwidth]{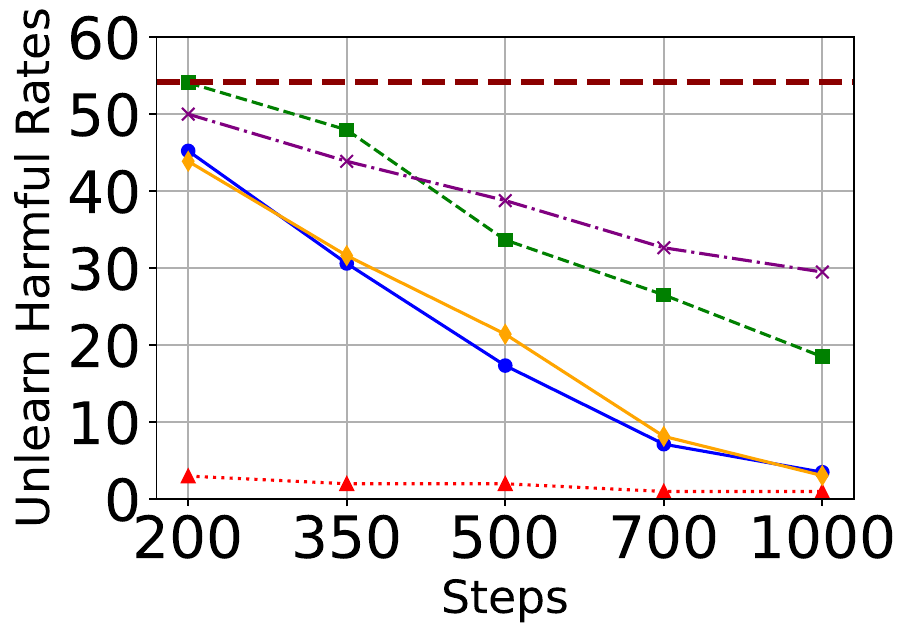}
        \subcaption{Harmful Rates vs Steps}
        \label{fig:appendix-opt2_7b_a}
        \end{subfigure}
        \begin{subfigure}{0.49\columnwidth}            
        \includegraphics[width=\columnwidth]{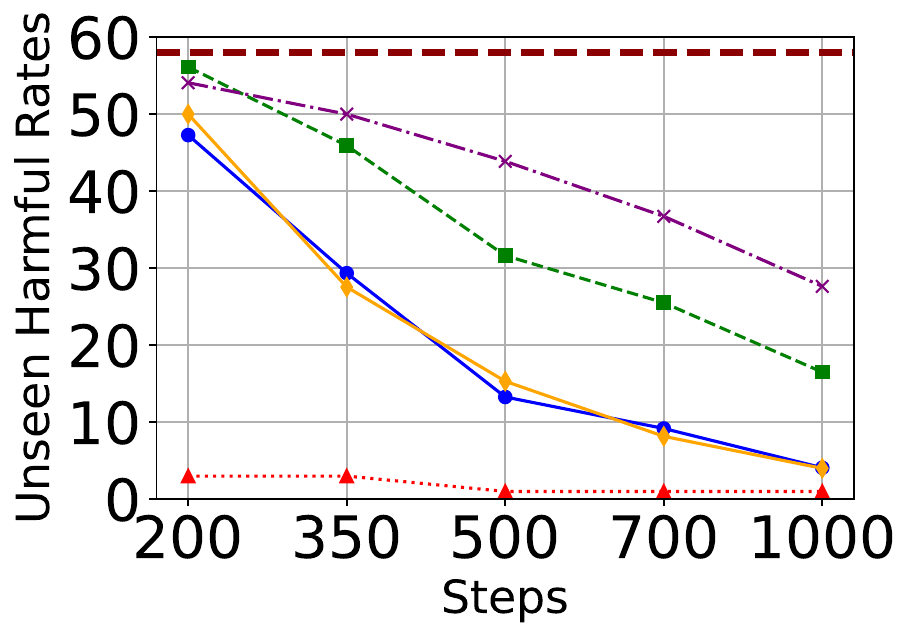}
        \subcaption{Harmful Rates vs Steps}
        \label{fig:appendix-opt2_7b_b}
        \end{subfigure}
        \begin{subfigure}{0.49\columnwidth}
        \includegraphics[width=\columnwidth]{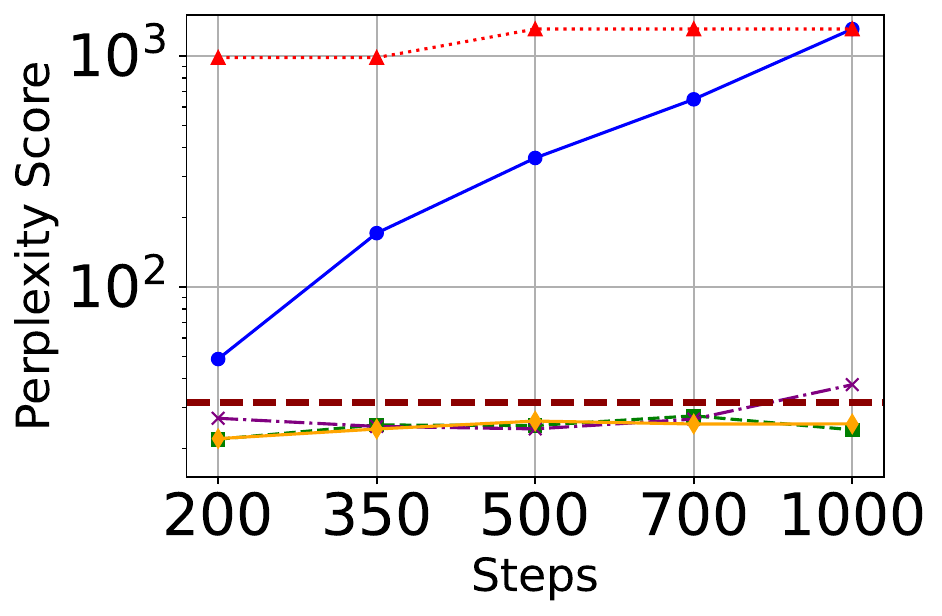}
        \subcaption{Perplexity Score vs Steps}
        \label{fig:appendix-opt2_7b_c}
        \end{subfigure}
        \begin{subfigure}{0.49\columnwidth}
        \includegraphics[width=\columnwidth]{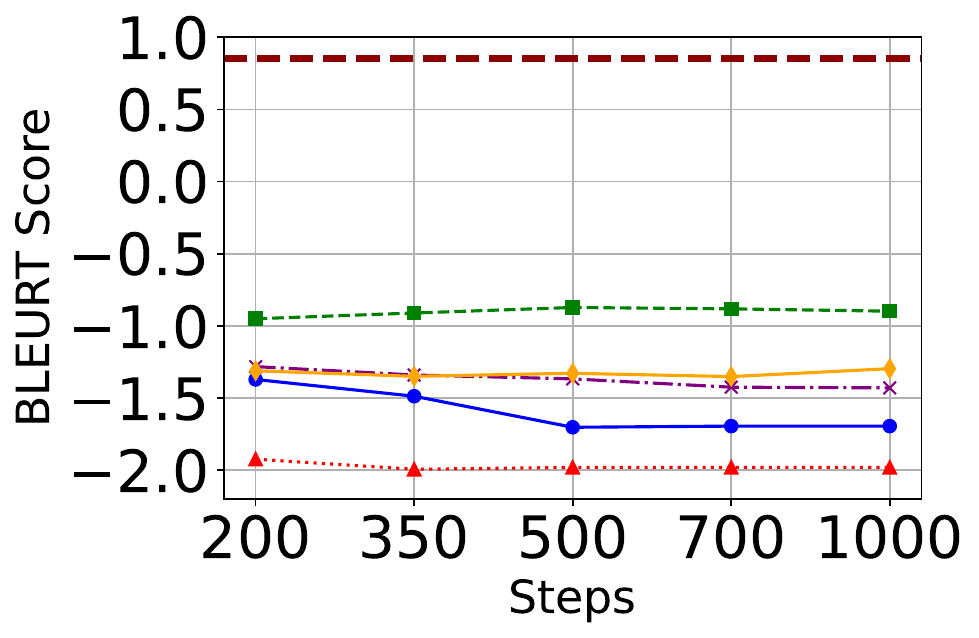}
        \subcaption{BLEURT Score vs Steps}
        \label{fig:appendix-opt2_7b_d}
        \end{subfigure}
\caption{
The performance of \method with a number of baselines on OPT-2.7B. Figure \ref{fig:appendix-opt2_7b_a} and Figure \ref{fig:appendix-opt2_7b_b} denotes the unlearning performance on unlearned and unseen samples, respectively. The $x$ axis represents the training steps and $y$ axis denotes the unlearn harmful rates. Figure \ref{fig:appendix-opt2_7b_c} and \ref{fig:appendix-opt2_7b_d} stands for the utility performance of each approach, where the $x$ axis represents the training steps and $y$ axis denotes the perplexity score and BLEURT score, respectively. The orange line represents the performance of \method.
}
\label{fig:appendix-opt2_7b}
\end{figure}

\begin{figure}
\centering
         \begin{subfigure}[b]{\columnwidth}
            \centering
            \includegraphics[width=0.9\columnwidth]{Figure/legend.jpg}
          \end{subfigure}
        \begin{subfigure}{0.49\columnwidth}
        \includegraphics[width=\columnwidth]{Figure/unlearned_Harmful.pdf}
        \subcaption{Harmful Rates vs Steps}
        \label{fig:appendix-llama_7b_a}
        \end{subfigure}
        \begin{subfigure}{0.49\columnwidth}            
        \includegraphics[width=\columnwidth]{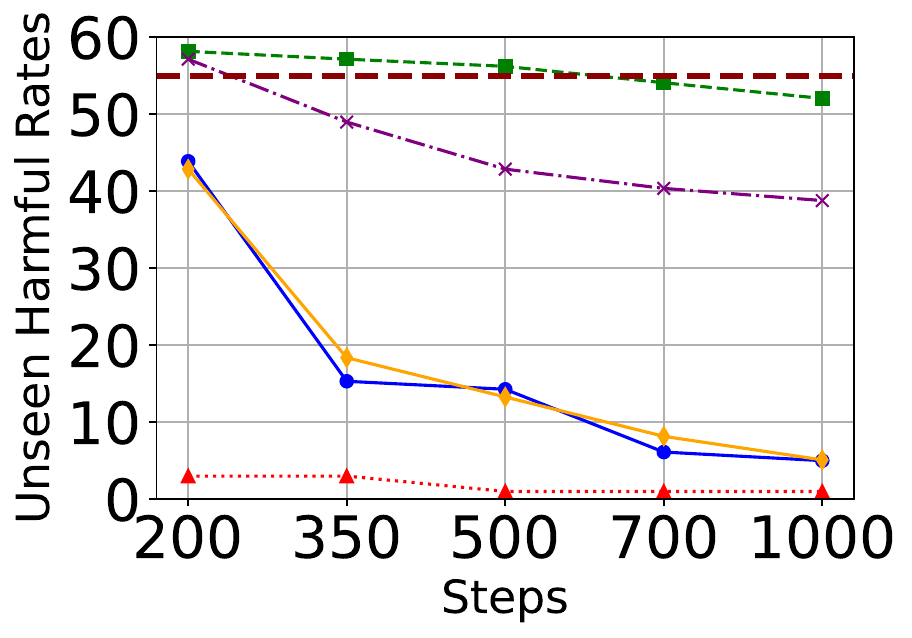}
        \subcaption{Harmful Rates vs Steps}
        \label{fig:appendix-llama_7b_b}
        \end{subfigure}
        \begin{subfigure}{0.49\columnwidth}
        \includegraphics[width=\columnwidth]{Figure/PPL.pdf}
        \subcaption{Perplexity Score vs Steps}
        \label{fig:appendix-llama_7b_c}
        \end{subfigure}
        \begin{subfigure}{0.49\columnwidth}
        \includegraphics[width=\columnwidth]{Figure/BLEURT.pdf}
        \subcaption{BLEURT Score vs Steps}
        \label{fig:appendix-llama_7b_d}
        \end{subfigure}
\caption{
The performance of \method with a number of baselines on LLAMA2-7B. Figure \ref{fig:appendix-llama_7b_a} and Figure \ref{fig:appendix-llama_7b_b} denotes the unlearning performance on unlearned and unseen samples, respectively. The $x$ axis represents the training steps and $y$ axis denotes the unlearn harmful rates. Figure \ref{fig:appendix-llama_7b_c} and \ref{fig:appendix-llama_7b_d} stands for the utility performance of each approach, where the $x$ axis represents the training steps and $y$ axis denotes the perplexity score and BLEURT score, respectively. The orange line represents the performance of \method.
}
\label{fig:appendix-llama2-7b}
\end{figure}

\begin{figure}
\centering
         \begin{subfigure}[b]{\columnwidth}
            \centering
            \includegraphics[width=0.9\columnwidth]{Figure/legend.jpg}
          \end{subfigure}
        \begin{subfigure}{0.49\columnwidth}
        \includegraphics[width=\columnwidth]{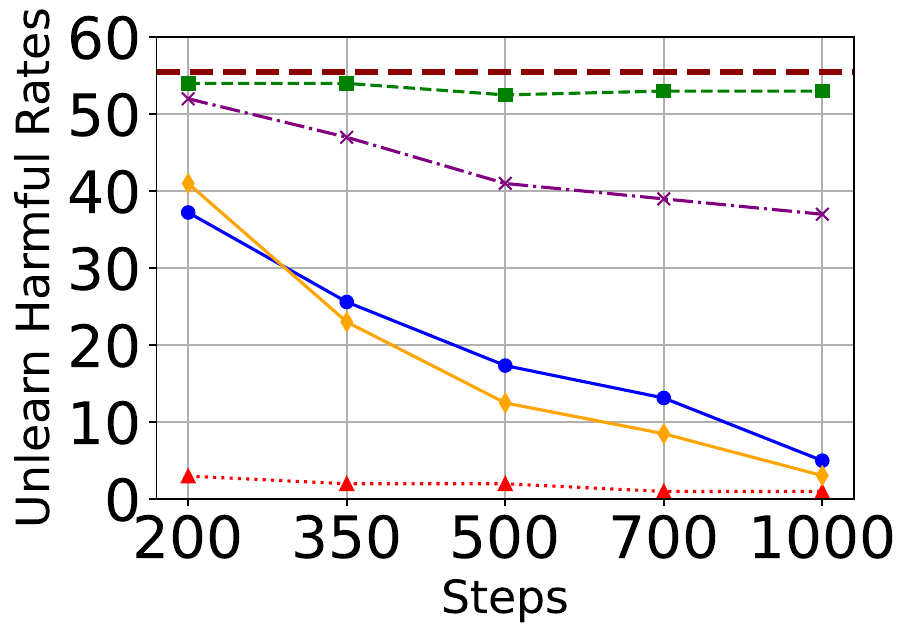}
        \subcaption{Harmful Rates vs Steps}
        \label{fig:appendix-llama_13b_a}
        \end{subfigure}
        \begin{subfigure}{0.49\columnwidth}            
        \includegraphics[width=\columnwidth]{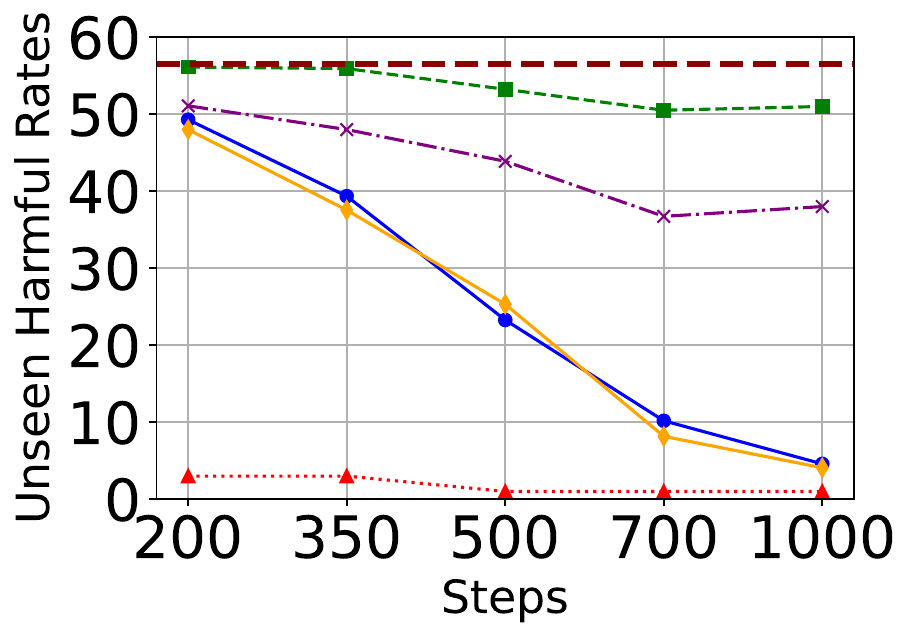}
        \subcaption{Harmful Rates vs Steps}
        \label{fig:appendix-llama_13b_b}
        \end{subfigure}
        \begin{subfigure}{0.49\columnwidth}
        \includegraphics[width=\columnwidth]{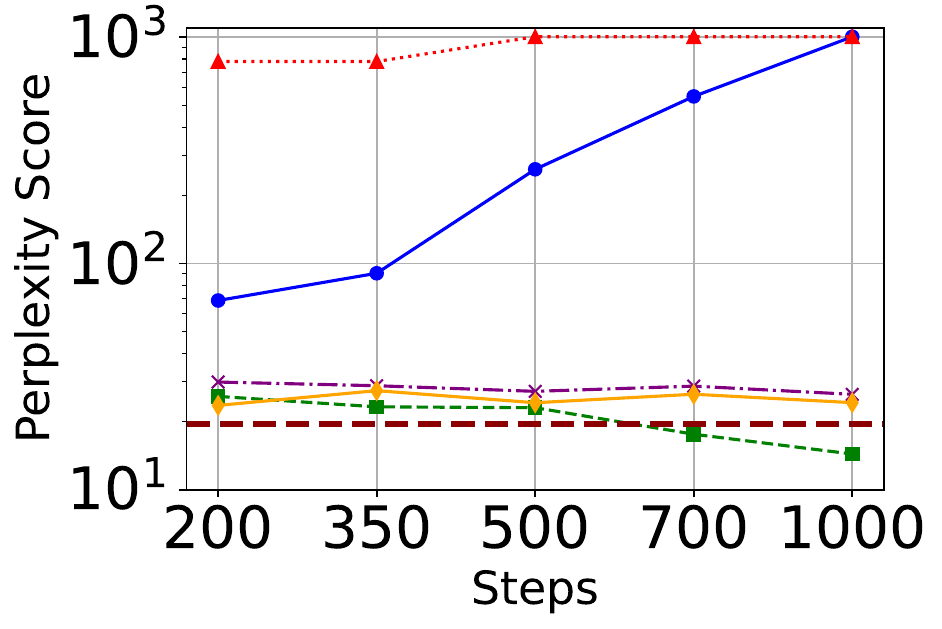}
        \subcaption{Perplexity Score vs Steps}
        \label{fig:appendix-llama_13b_c}
        \end{subfigure}
        \begin{subfigure}{0.49\columnwidth}
        \includegraphics[width=\columnwidth]{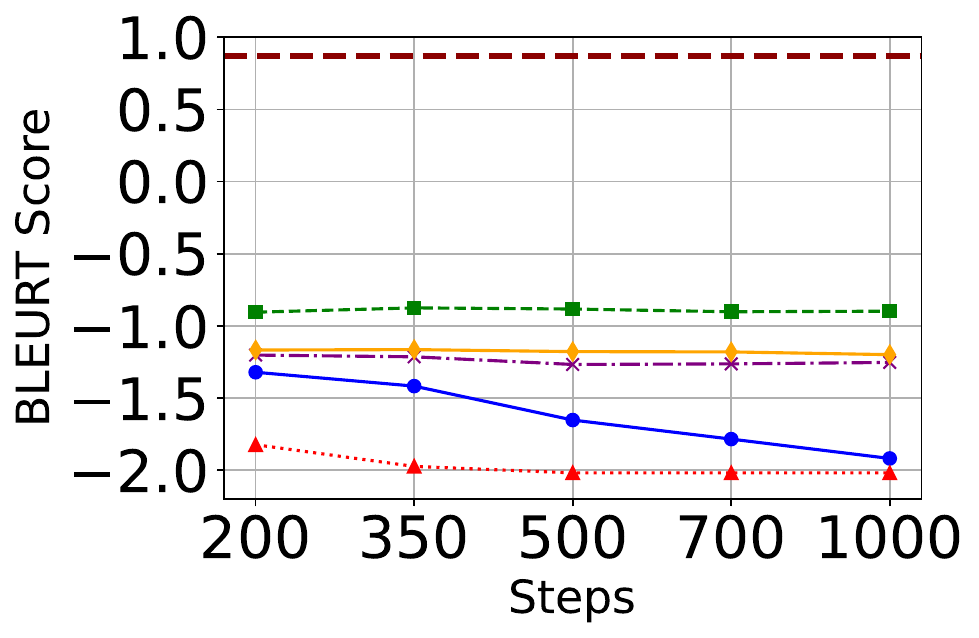}
        \subcaption{BLEURT Score vs Steps}
        \label{fig:appendix-llama_13b_d}
        \end{subfigure}
\caption{
The performance of \method with a number of baselines on LLAMA2-13B. Figure \ref{fig:appendix-llama_13b_a} and Figure \ref{fig:appendix-llama_13b_b} denotes the unlearning performance on unlearned and unseen samples, respectively. The $x$ axis represents the training steps and $y$ axis denotes the unlearn harmful rates. Figure \ref{fig:appendix-llama_13b_c} and \ref{fig:appendix-llama_13b_d} stands for the utility performance of each approach, where the $x$ axis represents the training steps and $y$ axis denotes the perplexity score and BLEURT score, respectively. The orange line represents the performance of \method.
}
\label{fig:appendix-llama2-13b}
\end{figure}

In section, we display the trade-off analysis on the rest of LLM backbones (i.e. OPT-2.7B, LLAMA2-7B (with unseen harmful rate) and LLAMA2-13B), shown in Figure~\ref{fig:appendix-opt2_7b}, Figure ~\ref{fig:appendix-llama2-7b} and Figure~\ref{fig:appendix-llama2-13b}, respectively. Similar to previous setup in Figure~\ref{fig:discussion_7b}, we show the performance of \method and the other baselines with different training steps. As demonstrated in the figures, throughout the training for all tested LLM architectures, \method consistently navigates the trade-off between unlearning and utility performance in a same trend as the previous setup.

\vspace{-0.2in}

\end{document}